\documentclass[review]{elsarticle}

\usepackage{hyperref}

\journal{}

\newcommand{\domain}[1]{\mathcal {#1}}
\begin{document}

\begin{frontmatter}

\title{A Short Review on Data Modelling for Vector Fields}

\author{Jun~Li, Wanrong~Hong, and Yusheng~Xiang}

\begin{abstract}
Machine learning methods based on statistical principles have proven highly successful in dealing with a wide variety of data analysis and analytics tasks. Traditional data models are mostly concerned with independent identically distributed data. The recent success of end-to-end modelling scheme using deep neural networks equipped with effective structures such as convolutional layers or skip connections allows the extension to more sophisticated and structured practical data, such as natural language, images, videos, etc. On the application side, vector fields are an extremely useful type of data in empirical sciences, as well as signal processing, e.g. non-parametric transformations of 3D point clouds using 3D vector fields, the modelling of the fluid flow in earth science and the modelling of physical fields. 

This review article is dedicated to recent computational tools of vector fields, including vector data representations, predictive model of spatial data, as well as applications in computer vision, signal processing, and empirical sciences.
\end{abstract}

\begin{keyword}
Vector fields, convolutional neural networks, spatial data analytics
\end{keyword}

\end{frontmatter}


\section{Introduction}
Vector fields are ubiquitous in vast scientific research fields. In a broad sense, vector fields arise when the object of interest has an underlying structure, which is often represented as a grid in a Euclidean space, and the variables at individual points make vectors of some physical or geometric significance.

The Euclidean structure of the data has been recognised for long \cite{Field1989, Bourlard1994} and well explored in the recent development of deep neural networks, for example, in the application areas such as image analytics \cite{Simoncelli2001}, and continuous temporal data analytics \cite{Graves2006}.

The successes of the applications can largely be attributed to the appropriate techniques for the exploitation of the Euclidean structure of the data domain. For example, the convolutional neural networks take advantage of the translational symmetry of the images defined on a two-dimensional plane \cite{LeCun1998}. Effective image analytics can be built upon the separation of deformation and scaling \cite{Bruna2013}.

There is an increasing tendency of applying the successful models to application scenarios where are the data arises from real-world physical processes across a spatial domain, including data from geoscience \cite{Li2004}, climate science\cite{Williams2016}, fluid dynamics \cite{Chen2007}, 3D geometry \cite{Zhao2011}, computer graphics \cite{Post2003}, to name a few.

However, most existing models treat the data as independent attributes across the domain without exploiting or accounting for the underlying significance of the vector fields. For example, in machine vision research, convolutional neural networks are highly successfully as models of the images and videos. The similar architectures are also straightforwardly used to represent the optical flow at pixels, as well as object deformations \cite{Fortun2015}. While it is reasonable to treat the red, blue and green channels across the domain of an image plane as independent attributes (scalar fields), the same approach may miss meaningful structures in the flow vector data over the domain.

\section{Background}

\subsection*{Learning-based Data Models}
{\em Statistical learning} is a task to discover and formally represent the relationship between two parts of information. The relationship is often formulated in some functional form $\mathcal X \mapsto \mathcal Y$. The information in the domain of $\mathcal X$ is always observable and usually denoted as {\em data attributes} or {\em input variables}. The information in the domain $\mathcal Y$ represents the target of interest. 
In {\em supervised learning}, the joint examples of $X \in \mathcal X$ and $Y \in \mathcal Y$ are available and used to figure out the configuration of a data model in the {\em training} process. Beyond the training examples, the variables of $\mathcal Y$ is not directly available.

In theory, any function approximators can be used for statistical learning. Practically, in recent years deep neural networks has achieved excellent results in a wide range of application areas. 

In particular neural networks with convolutional design and multistage architecture (deep), can leverage statistical properties of the data, such as stationarity and compositionality in the $\mathcal X$ domain. This makes them effective in dealing with data with spatial distributions including the field data that is the focus of this review.



\subsection*{Data Models of Multiple Targets}


Vector field data analysis involves modelling multiple responses at the same spatial location. From the modelling's point-of-view this is related to the
{\em multiple task learning} ({\em MTL}), in which multiple learning tasks are solved jointly. The information can be shared across some or all of the tasks. So in the face of data insufficient problem. MTL can usually achieve better results than that is learned via individual tasks. 



\subsection*{Data Models of Field Data}
The study of fields deals with functions defined on a domain of Euclidean metric. Most fields are abstract mathematical functions, that map points in spatial domains to ranges. The dimensionality of a spatial domain is uncertain. It varies from one-dimensional to n-dimensional. The range can be a scalar or a vector of two or more dimensions. Hence, fields are divided into scalar and vector fields.

In the engineering world, a practical instance consists of values sampled at regular grids in the domain. For example in computer vision, 3-D shapes can be represented as grids by constructing local geometric descriptors, which capture properties such as curvature. In voice analysis, two words can be connected if they often appear near each other in the co-occurrence graph.

\subsection*{Data Models of Vector Fields}

Vector fields arise when we simultaneously consider a spatially distributed object with multiple responses at each location. I.e. in a vector field, the dimension of the data has two folds of complexity, the measurements at individual locations are multidimensional and there are many measurements across a spatial domain. 

Like the case of scalar fields, different types of tasks are defined on vector fields, i.e. model the fields, regional and global characteristics. The vector form allows defining extra geometrical physical characteristics notions, such as the {\em flux} and the {\em divergence}, which do not present in scalar fields treated individually.




\section{Methodology}

From the viewpoint of machine learning, i.e. to build a function to approximate the relationship between different aspects in some process of interest, a vector field model is of two characteristics. On the ``vector'' side, multiple outputs are considered for one input simultaneously. On the ``field'' side, it is needed to consider the outputs on multiple inputs simultaneously to exploit the relationship between inputs in the domain of the function. 

In this section, we discuss related techniques to model vector field data. We mainly organise the materials using the taxonomy discussed above by first considering the techniques addressing multiple outputs in learning, followed by field data models and then works in the intersection of both areas. The perspective is mainly from how the problem is defined and approached. It is worth noting that an orthogonal viewpoint of no less importance is via the computing techniques, such as matrix factorisation, regularised optimisation as well as neural network training. Our choice of taxonomy is partially for the convenience of discussion and partially to help understand the interdisciplinary relationship of works of various origins. There is a brief summarisation of different computational techniques adopted in various modelling strategies.

\subsection {Modelling Multiple Targets}
We start with reviewing the research of multiple task learning. In the artificial intelligence community, it has been long to notice that simultaneously learning multiple tasks can have a synergistic effect, where the relationship between the input and the output of each task is better modelled than to learn individual tasks independently. Extra tasks can be introduced to represent additional knowledge about the desired task. For example, in an early study \cite{AbuMostafa1990}, the prior knowledge of the target function to be invariant to irrelevant transforms of the data has been represented using artificial examples. In \cite{Baxter1995}, it has been shown that to learn a common representation of the data to perform multiple tasks can help improve sample efficiency in supervised learning, in terms of the rate at which generalisation risk reduces with the growth of training examples. 

Roughly, the two early examples represent two ways of exploiting and dealing with the relationship between the multiple tasks: considering the relationship between tasks in the functional spaces and discovering a common data representation to facilitate all tasks. More formally, the former one is concerned about the information exchange between $\domain{F}^m$, where $\domain{F}: \domain{X}\mapsto \domain {Y}$ and $m$ is the number of tasks. An $f \in \domain{F}$ is a target function. To be explicit, $f_j: \domain X_j \mapsto \domain Y_j, j \in \{1\dots m\}$. Note $\domain X_i$ and $\domain X_j$ can refer to the same space when $i\neq j$, so can $\domain Y_i$ and $\domain Y_j$. The subscript emphasises different tasks. 

The latter task, on the other hand, explicitly models $\domain X_1 \mapsto \domain Z, \domain X_2 \mapsto \domain Z, \dots$ and $\domain Z \mapsto \domain Y_1, \domain Z \mapsto \domain Y_2, \dots$ where Z is shared over all tasks. Finer taxonomy exists in related literature of multiple task learning \cite{Zhang2018}, but for our interest, as a component of modelling vector fields, the rough classification would suffice.

\subsection{Functional Relationship}
For the convenience of discussion let us assume that different tasks share the same target domain, i.e. $\domain Y_1 = \domain Y_2 = \dots = \domain Y$. So all functions of interest belongs to $\domain F: \domain X \mapsto \domain Y$. Without loss of generality for practical applications, we consider square integrable functions, $\int_x f(x)^2 < \infty$, between which some (dis-) similarity measure can be defined $0 \leq D(f1, f2) < \infty$. 

Since the tasks are related, the fundamental idea is that the different task functions are similar to some extent \cite{BenDavid2003}. The similarity can be formulated into different forms of regularisation to assist the learning. The structure between different tasks has been studied early in \cite{Thrun1996}, where distance metrics are learned for task-specific nearest-neighbour classifiers. The metrics are transferred among tasks to represent a  shared hypothesis space. A Bayesian treatment has been proposed in \cite{Crammer2012}.

\cite{Bakker2003} built a Bayesian model of the relationship between multiple tasks and showed the empirical advantage of grouping the task and performed simultaneous learning. \cite{Micchelli2005} studied the relationship between the elements of the output vectors (tasks) under a vector-valued reproducing kernel Hilbert space theoretical framework. \cite{Evgeniou2005} developed a kernel-based multi-task learning algorithm based on regularisation methods and demonstrated empirical advantages.

\subsection{Subspace representation learning for Multiple Tasks}
It has long been realised that for most practical machine learning problems, a model of multiple computational stages tends to reveal a latent relationship in the data and achieve better performance than direct modelling. The intermediate competition results make a new form of representation of the data, which can be adapted to extract information for performing subsequent tasks efficiently. Multiple related tasks can share common intermediate data representation. The supervision from one task to help improve the representation of data for other tasks.

Shared intermediate representation for multi-task learning was introduced in an early multi-layer neural network, where the hidden layer was shared among different tasks and served as the common representation \cite{Caruana1997}. Further developments have introduced carefully designed hidden layer structures \cite{Liao2005} and mechanism of describing task-specific context \cite{Silver2008}. 

Another representative of the multi-task feature learning framework has been proposed based on discovering a linear transformation of the raw attributes of the data to form a common basis of features for efficient representation. Efficient computational methods has been proposed based on (convex) regularisation optimisation \cite{Argyriou2007,Argyriou2008}.

Deep neural networks have shown promise to discover the effective representation of the data for various tasks. Research works on developing deep architectures for multiple tasks, which can learn more effective data representation than networks trained individually. Multi-task neural networks have been shown useful in machine vision and image processing tasks, including facial keypoints detection \cite{Zhang2014}, video summarisation \cite{Liu2015}, biological-image analysis \cite{Zhang2016} as well as human pose estimation. Multi-task learning has also shown advantages in neural networks applied to natural language processing \cite{Mrksic2015,Wen2017,Liu2017}.

\section{Applications}
Vector field data modelling has been applied to a wide range of applications. As discussed in the previous sections, the techniques dealing with vector fields are special cases of statistical learning from data. The traditional application areas of machine learning models include dealing with multimedia signals, such as images videos or readings from other sensors. Vector field data modelling has been applied in the analysis of optical flow in machine vision, and recently to 3D geometry data. In this section, we begin with applications of a vector field modelling in computer vision and the general three-dimensional geometry learning, followed by the applications in other scientific areas.

\subsection{Image and Video Analytics}
As the description of the dynamics between two subsequently observed images, the optical flow has essential applications in machine vision, such as motion estimation, fore-/background segmentation, action analysis, crowd analysis, \cite{Lucas1981,Fortun2015} etc.

\cite{Modlin2012} modelled vector fields in a polar coordinate system by considering circular statistics. The statistical properties of the vector fields are studied and employed to model optical flow in \cite{Roth2005}. In \cite{Li2013}, a Bayesian factorisation model has been proposed for vector fields and tested for image optical flow. Recently, with the advance of the deep neural networks. There has been increasing interest in modelling optical flow vector fields directly from a pair of images (the end-to-end paradigm) using deep convolutional neural networks(CNNs) based models.

In an early study \cite{Dosovitskiy2015}, FlowNet applied CNNs to optical flow prediction for the first time. Although it has a fast computation speed, its performance is still not as good as the best traditional method concerning the quality of the flow, mostly limiting the application of FlowNet. Flownet2.0\cite{Ilg2016} was advanced based on FlowNet, by adjusting dataset schedules and introducing a stack architecture, achieving impressive results. In \cite{Ranjan2017}, the authors combine the classic spatial-pyramid formulation with CNNs to compute optical flow. Recently, PWC-Net\cite{Sun2018} performs compact but effective in the optical flow since it is based on three basic principles: pyramidal processing, warping and cost volume. \cite{Yin2018} introduces an unsupervised learning network, called GeoNet, for analyzing video by jointly learning monocular depth, optical flow and egomotion estimation. Furthermore, optical flow also has critical applications in medical images; for example, in \cite{Balakrishnan2018}, a learning-based framework is designed for deformable medical image registration. Instead of learning vectors element-by-element, the framework in \cite{Vos2018} learns to predict transformation parameters, which then can directly be used to form a dense displacement vector field.

\subsection{Point Cloud Modelling}
The natural transition of our need from 2D graphic computation towards 3D point set processing technology results from the new technologies to acquire 3D point clouds and developments in other fields. 3D techniques are widely used in various areas, such as 3D prints, computer games, architecture construction, automotive design, archaeology, biology, and art. By building the vector field at all points, different 3D processing, including segmentation, registration, surface reconstruction, and compression, can be accessed. For single-view 3D object reconstruction, in \cite{Li2018}, a 3D voxel representation based on vector field is introduced to address the limitation, the waste of empty voxels and background pixels, from exiting CNN-based methods. 3D point cloud registration is to align two point clouds by applying a spatial transformation. In an early study\cite{Tubic2002}, the vector field was firstly introduced to 3D data representation for point cloud processing.  In \cite{Nguyen2013}, Nguyen et al. introduce vector field representation for 3D point clouds into an automatic coarse-to-fine rigid registration method, which avoids the computation of searching for closest points, compared to the classic Iterative Closest Point(ICP) algorithm. Xu et al.\cite{Xu2015} designed an out-of-core method for surface reconstruction is another application of the vector field in point clouds. According to \cite{Xu2015}, in the vector field, each vector serves as the direction and the distance between a point and the closest point on the surface to be reconstructed.

\subsection{Fundamental Sciences}
Traditionally, machine learning is applied to areas where there is no explicit expression of the relationship between the variables of interest. When such a relationship is available, the problem is often deemed subject to conventional scientific investigation. However, it has been increasingly realised that in a learning process, the machine is required to mimic a cognitive system to respond or to identify targets of interest in complex observations of the environment. The paradigm mirrors the process of performing scientific exploration to discover governing natural laws using empirical observations and experiments. There is no essential distinction between such a data-driven approach and the investigation of the fundamental sciences, despite the seemingly different approaches to perform analytics of the observed phenomena. Field data is ubiquitous to characterise real-world phenomena or processes. We will briefly review the advances in adopting machine learning methods for field data modelling in a few scientific areas to point out this promising new direction of the investigation.

\subsubsection{Materials science}

Data-driven discovery in material science is a burgeoning area recently. Learning the relationship between the chemical composition and the properties of the material happens in different levels of granularity \cite{Ramprasad2017}, including gross property prediction \cite{Ghiringhelli2014}, molecular fragment level characteristics \cite{vanKrevelen2012} and atomic level analytics \cite{Chmiela2017}. Atomic-level analytics is concerned with the effect of the spatial distribution of atoms in chemical composition. 

In \cite{Behler2014}, neural networks have been employed to represent the potential energy function given the atomic configuration of some unknown material. It is necessary to include some critical spatial invariance in the representation, including translation, rotation and atomic-level symmetry characteristics \cite{Behler2007}.

\subsubsection{Earth science} 

Geographic vector fields are the subject of interest in various earth sciences \cite{Liu2018, Deser2010, Li2004}, where a systematic field data modelling framework has been established \cite{Tomlin1990}.

In \cite{Hodgson1996}, the local characteristics of a vector field, such as the vector mean and dispersion, have been discussed and modeled. The spatial model of the vector field can be used for subsequent tasks, such as understanding the plant seed dispersion process \cite{Qiu2008} or predicting microscale land activities \cite{Marquinez2003}.

To a larger scale, the focal analytics studies the neighbours of a vector field, such as the surface slope or angles \cite{Hodgson1998}. The analytics is particularly useful in modelling terrains \cite{Corripio2003, Hengl2009}.

The global scale modelling of the field data can be used to discover statistically significant signals in multiple or historical observations \cite{Klink1989, Kossin2010, Church2013}.

The recent success of using deep neural networks-based modelling for image and multimedia data has been introduced to and shown promise in physical field modelling. In \cite{Gagne2020}, the adversarial generative neural network is adopted to characterise the chaotic dynamics in the Lawrenz '96 model. Ham et al. \cite{Ham2019}, has constructed a convolutional neural network to predict the El Niño/Southern Oscillation effect using the field data of sea surface temperature.

\section{Conclusion}
In this short review article, we have introduced recent important computational models for learning to model vector field data, including the data analysis models and predictive data analytics tasks. We have also reviewed a few important application areas of vector field data models in multimedia signal processing and empirical sciences. Some open research problems on multi-task learning have been discussed as well.

\bibliography{Literature}
\end{document}